\documentclass[journal,twoside,web]{ieeecolor}
\usepackage{generic}
\usepackage{amsmath,amssymb,amsfonts}
\usepackage{algorithm,algorithmic}
\usepackage{graphicx}
\usepackage{textcomp}
\usepackage[table]{xcolor}
\usepackage{booktabs}
\usepackage{array}
\usepackage{multirow}
\usepackage{setspace}
\usepackage{caption}
\usepackage{xcolor}
\usepackage{soul}
\sethlcolor{yellow}
\soulregister\ref7
\soulregister\cite7
\soulregister\pageref7
\soulregister\eqref7

\usepackage{float}
\usepackage{textgreek}
\usepackage{mathptmx}
\usepackage[T1]{fontenc}
\usepackage{listings}
\let\labelindent\relax
\usepackage{enumitem}
\usepackage{hyperref}
\hypersetup{hidelinks=true}
\usepackage[style=ieee, backend=biber]{biblatex} 

\definecolor{codegreen}{rgb}{0,0.6,0}
\definecolor{codegray}{rgb}{0.5,0.5,0.5}
\definecolor{codepurple}{rgb}{0.58,0,0.82}
\definecolor{backcolour}{rgb}{0.95,0.95,0.92}

\setlist[itemize]{leftmargin=*}
\setlist[enumerate]{leftmargin=*}

\lstdefinestyle{mystyle}{
  backgroundcolor=\color{backcolour},  
  numberstyle=\tiny\color{codegray},
  basicstyle=\ttfamily\footnotesize,
  breakatwhitespace=false,         
  breaklines=true,                 
  captionpos=b,                    
  keepspaces=true,                 
  numbers=left,                    
  numbersep=5pt,                  
  showspaces=false,                
  showstringspaces=false,
  showtabs=false,                  
  tabsize=2,
  stringstyle=\ttfamily,
  commentstyle=\ttfamily,
  morecomment=[s]{"""}{"""},
  morestring=[s]{"""}{"""}
}

\def\BibTeX{{\rm B\kern-.05em{\sc i\kern-.025em b}\kern-.08em
    T\kern-.1667em\lower.7ex\hbox{E}\kern-.125emX}}
\begin{document}

\title{Patients Speak, AI Listens: LLM-based Analysis of Online Reviews Uncovers Key Drivers for Urgent Care Satisfaction}

\author{
    Xiaoran Xu$^{*}$,
    Zhaoqian Xue$^{*}$,
    Chi Zhang,
    Jhonatan Medri,
    Junjie Xiong,
    Jiayan Zhou,
    Jin Jin,
    Yongfeng Zhang,
    Siyuan Ma$^{\dagger}$,
    and Lingyao Li$^{\dagger}$
    \thanks{*These authors contributed equally to this work.}
    \thanks{${\dagger}${Corresponding authors.}}
    \thanks{Xiaoran Xu is with the Electrical Engineering department, University of South Florida, Tampa, United States (e-mail: xiaoranxu@usf.edu).}
    \thanks{Zhaoqian Xue is with the Department of Biostatistics, Epidemiology and Bioinformatics, University of Pennsylvania, Philadelphia, United States (e-mail: Zhaoqian.Xue@PennMedicine.upenn.edu).}
    \thanks{Chi Zhang is with the Computer Science and Engineering department, University of South Florida, Tampa, United States (e-mail: chiz@usf.edu).}
    \thanks{Jhonatan Medri is with the Mathematics \& Statistics, University of South Florida, Tampa, United States (e-mail: jm192@usf.edu).}
    \thanks{Junjie Xiong is with the Department of Computer Science and Engineering, University of Missouri Science and Technology, Rolla, United States (e-mail: junjiexiong@mst.edu).}
    \thanks{Jiayan Zhou is with the School of Medicine, Stanford University, Stanford, United States (e-mail: jyzhou@stanford.edu).}
    \thanks{Jin Jin is with the Department of Biostatistics, Epidemiology and Bioinformatics, University of Pennsylvania, Philadelphia, United States (e-mail: Jin.Jin@Pennmedicine.upenn.edu).}
    \thanks{Yongfeng Zhang is with the Department of Computer Science, Rutgers University, New Brunswick, United States (e-mail: yongfeng.zhang@rutgers.edu).}
    \thanks{Siyuan Ma is with the Department of Biostatistics, Vanderbilt University, Nashville, United States (e-mail: siyuan.ma@vumc.org).}
    \thanks{Lingyao Li is with the School of Information, University of South Florida, Tampa, United States (e-mail: lingyaol@usf.edu).}
}

\maketitle

\begin{abstract}
Investigating the public experience of urgent care facilities is essential for promoting community healthcare development. Traditional survey methods often fall short due to limited scope, time, and spatial coverage. Crowdsourcing through online reviews or social media offers a valuable approach to gaining such insights. With recent advancements in large language models (LLMs), extracting nuanced perceptions from reviews has become feasible. This study collects Google Maps reviews across the DMV and Florida areas and conducts prompt engineering with the GPT model to analyze the aspect-based sentiment of urgent care. We first analyze the geospatial patterns of various aspects, including interpersonal factors, operational efficiency, technical quality, finances, and facilities. Next, we determine Census Block Group (CBG)-level characteristics underpinning differences in public perception, including population density, median income, GINI Index, rent-to-income ratio, household below poverty rate, no insurance rate, and unemployment rate. Our results show that interpersonal factors and operational efficiency emerge as the strongest determinants of patient satisfaction in urgent care, while technical quality, finances, and facilities show no significant independent effects when adjusted for in multivariate models. Among socioeconomic and demographic factors, only population density demonstrates a significant but modest association with patient ratings, while the remaining factors exhibit no significant correlations. Overall, this study highlights the potential of crowdsourcing to uncover the key factors that matter to residents and provide valuable insights for stakeholders to improve public satisfaction with urgent care.
\end{abstract}


\section{Introduction}

Urgent care plays a key role in the healthcare system by offering quick, accessible treatment for non-life-threatening conditions. It provides a more efficient and cost-effective option for minor injuries and sudden illnesses compared to primary care, which focuses on long-term health, and emergency rooms, which handle severe cases but often involve higher costs and longer waits \cite{weinick2009urgent}. Therefore, ensuring timely access to urgent care facilities is crucial for maintaining public health and enhancing community well-being, particularly as urban populations continue to grow. Customer satisfaction in urgent care is particularly important as it reflects the quality, responsiveness, and efficiency of the care provided, directly influencing patient outcomes, trust in healthcare providers, and decisions regarding future healthcare utilization \cite{qin2014ucperf,knowles2012patients}. 

Conventional approaches to measuring patient satisfaction and experiences with healthcare services \cite{al2014patient,howard2007patient} are often based on surveys and interviews, such as the Household Pulse Survey (HPS) conducted during the COVID-19 pandemic \cite{uscensus2024}. However, these surveys and interviews are typically unable to fully capture the nuances and diversity of residents’ experiences due to limitations in time, spatial coverage, and resource intensity. For example, large-scale surveys like HPS, though thorough, can be resource-intensive and time-consuming. Such limitations in conventional methods can restrict the ability of healthcare providers and policymakers to promptly understand and respond to the evolving needs of the local community.

Recent research in crowdsourcing methodologies, especially the ones using social media and online review platforms, has provided great opportunities for us to understand healthcare services due to the richness of user-generated content \cite{wang2020crowdsourcing,li2022dynamic}. Current studies have explored the potential of crowdsourcing in multiple facets, including evaluating customer experiences related to healthcare services \cite{bentan2024addressing}. Crowdsourcing via social media or online reviews can capture diverse voices and continuous feedback from customers, providing a more representative community sentiment about urgent care services and highlighting aspects that residents value most.

However, analyzing crowdsourced data poses unique challenges due to the unstructured format, varying text length, and differential language use across diverse sources. Traditional natural language processing (NLP) approaches often fall short of accurately interpreting such data, as they typically rely on coarse-grained sentiment labels, struggle to scale to large and continuously evolving review datasets, and provide limited interpretability for identifying actionable service dimensions in healthcare contexts. Recent advancements in large language models (LLMs) have illustrated the capability to effectively derive insights from large-scale textual data~\cite{li2024scoping,gao2024examining}. For instance, researchers have successfully utilized LLMs to extract detailed clinical information from electronic health records (EHRs)~\cite{yang2022large} and identify sentiment and key themes from patient reviews of medical treatments~\cite{kornblith2025analyzing}. Additionally, current studies have leveraged LLMs to monitor public discourse~\cite{espinosa2024use}, derive consumer insights~\cite{matheny2024enhancing}, and predict health decisions~\cite{ding2024leveraging}. These applications highlight the potential of LLMs in transforming diverse, unstructured data sources into valuable knowledge that can support healthcare decision-making.

Despite these advances, there are still important gaps in understanding patient satisfaction with urgent care services. Existing healthcare sentiment analysis studies largely focus on overall sentiment or thematic trends, often using traditional machine learning methods, without disentangling specific service dimensions that shape patient experiences~\cite{al2014patient,howard2007patient,shah2021mining}. Recent LLM-based studies demonstrate strong capability in extracting insights from unstructured healthcare text~\cite{li2024scoping,gao2024examining,yang2022large}, yet few have systematically applied aspect-based sentiment analysis tailored to urgent care settings, nor examined how fine-grained service perceptions interact with area-level socioeconomic context. As a result, the determinants of urgent care satisfaction and their robustness across diverse communities remain insufficiently understood. To address these gaps, this study examines patient satisfaction through two primary research questions:

\begin{itemize}
\item \textbf{RQ1:} What patterns emerge across distinct sentiment dimensions in urgent care services, as identified by state-of-the-art LLM analytics?
\item \textbf{RQ2:} Which of these sentiment dimensions most strongly drive overall patient satisfaction, and do these effects persist independent of the facilities' area-level socioeconomic characteristics?
\end{itemize}

This study leverages large-scale Google Maps healthcare reviews and state-of-the-art LLM-based aspect-based sentiment analysis to examine urgent care satisfaction across diverse geographical areas. By integrating fine-grained sentiment dimensions with area-level socioeconomic characteristics, we identify the key service aspects that drive patient satisfaction and assess whether these effects persist across different community contexts. Our findings provide actionable insights for healthcare providers and policymakers to improve urgent care delivery and promote equitable access to high-quality services.

\section{Data and Methods}

\subsection{Data Preparation} 
\begin{figure*}[htbp]
\centering
\includegraphics[width=\textwidth]{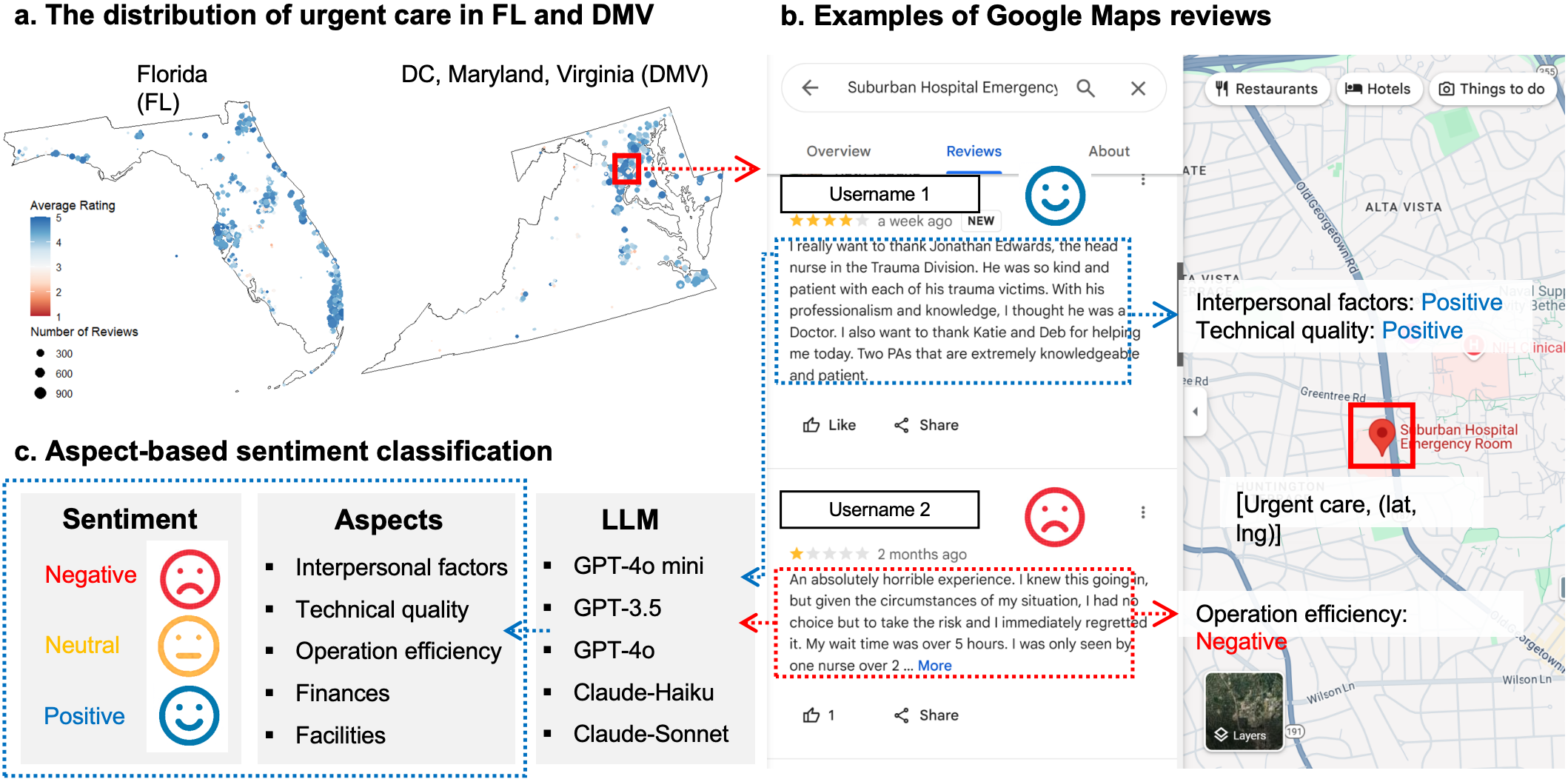}
\caption{Workflow of data processing and analysis of Google Maps
reviews. \textbf{a.} Distribution of urgent care facilities in Florida and
DMV (the District of Columbia, Maryland, and Virginia). Each point
represents an urgent care facility, with the point size indicating the
volume of reviews. \textbf{b.} Examples of patient reviews, one negative and
one positive. \textbf{c.} LLMs used to classify opinions extracted from
Google Maps reviews.}
\label{fig:workflow}
\end{figure*}

Our framework for analyzing urgent care sentiment from Google Maps reviews is summarized in Figure \ref{fig:workflow}. In this study, Google Maps reviews constitute our primary data source for two principal reasons. First, Google Maps has witnessed a substantial increase in review volume, surpassing other platforms such as Yelp and TripAdvisor \cite{munawir2019visitor}. Second, unlike social media platforms such as Facebook or Twitter, Google Maps provides targeted insights directly related to service quality \cite{yum2024enhancing}. We utilize the Google Maps review dataset \cite{li2022uctopic,yan2023personalized} published by the University of California, San Diego, covering reviews through September 2021. The dataset comprises (1) patient review data (e.g., username, rating, review text, and timestamp), and (2) POI metadata (e.g., address, geographical coordinates, categories, average ratings, total number of ratings).

Next, we conduct several steps to filter the dataset. We use the keyword ``urgent care'' as the POI filtering condition to select targeted reviews, ensuring their direct relevance to urgent care facilities. We then select the DMV area and Florida as case studies because both regions feature diverse populations and numerous urgent care centers, enabling comprehensive analysis of patient experiences across different demographic and geographic contexts. The resulting dataset contains 250,588 reviews covering 1,225 healthcare facilities providing urgent care services. To specifically extract reviews reflecting residents' attitudes toward urgent care, we exclude entries containing only numerical ratings without textual reviews. This yields 152,680 reviews associated with users’ feedback toward urgent care facilities. 

\subsection{Patient Satisfaction Analysis Framework}
To effectively categorize patient experiences expressed in Google Maps reviews, the first step is to identify the key aspects associated with patient reviews. Our analysis draws insights from established medical service quality frameworks, as summarized in Table \ref{tab:healthcare-quality}.

Based on the studies presented in Table \ref{tab:healthcare-quality}, we first recognize that certain aspects, such as prescription-related services, provider function, and touch point diversity, are often not explicitly captured in Google Maps review data. Meanwhile, aspects such as comfort, communication, interpersonal manner, empathy, and responsiveness often exhibit conceptual overlap, which presents challenges for language models in distinguishing fine-grained aspect-based sentiments. To address these limitations, we consolidate related attributes into broader categories, for example, merging elements such as wait time, process efficiency, and managerial decisions under operational efficiency. Therefore, we identify five key dimensions for the framework: interpersonal interactions, technical quality, financial considerations, operational efficiency, and facility conditions. These dimensions offer a comprehensive yet concise representation of patient experiences in urgent care settings. Furthermore, they align with well-established healthcare quality assessment frameworks (see Table \ref{tab:healthcare-quality}) and provide an appropriate level of granularity to support effective aspect-based sentiment analysis using LLMs.

\begin{itemize}
\item \textbf{Interpersonal factors:} Features that quantify how well providers interact personally with patients (e.g., concern, friendliness, courtesy, disrespect, rudeness).
\item \textbf{Technical Quality:} Competence of providers and adherence to high standards of diagnosis and treatment (e.g., thoroughness, accuracy, unnecessary risks, mistakes).
\item \textbf{Operational Efficiency:} The results of medical care encounters (e.g., helpfulness of medical care providers in improving or maintaining health).
\item \textbf{Finances:} Factors involved in paying for medical services (e.g., reasonable costs, alternative payment arrangements, comprehensiveness of insurance coverage).
\item \textbf{Facilities:} Presence of medical care resources (e.g., availability of healthcare facilities and equipment)
\end{itemize}

\begin{table}[htbp]
\centering
\caption{Consolidation of Healthcare Quality Aspects from Existing Studies.}
\label{tab:healthcare-quality}
\begin{tabular}{p{2.5cm}p{5.5cm}} 
\hline
\textbf{Author \& Year} & \textbf{Aspects} \\
\hline
Ware et al. (1983)\cite{ware1983defining} & Interpersonal manner, technical quality, efficacy/outcomes, accessibility/convenience, finances, continuity, physical environment, facilities/availability \\
\hline
Zeithaml et al. (1990)\cite{zeithaml1990delivering} & Reliability, assurance, tangibles, empathy, responsiveness. \\
\hline
Marshall and Hays (1994)\cite{marshall1994patient} & Access to care, financial aspects, availability of services, continuity of care, technical quality, interpersonal care \\
\hline
Tucker and Adams (2001)\cite{tucker2001incorporating} & Caring, empathy, reliability, responsiveness, access, communication, outcomes \\
\hline
World Health Organization. (2006)\cite{who2006quality} & Effectiveness, efficiency, accessibility, patient-centeredness, safety, equity. \\
\hline
Manary et al. (2013)\cite{manary2013patient} & Communication with nurses \& physicians, timeliness of assistance, Pain management, discharge planning, cleanliness \& facilities, emotional support. \\
\hline
Ozcelik et al. (2019)\cite{ozcelik2019customer} & Provider type \& function, touchpoint diversity, patient interaction, environment, psychological support, process efficiency, financial accessibility. \\
\hline
Agarwal et al. (2019)\cite{agarwal2019online} & Comfort, professionalism, facilities, pediatric care, staff interactions, communication, waiting, billing/insurance, pain management, diagnostic testing, prescription and pharmacy, reception. \\
\hline
\end{tabular}
\end{table}

\subsection{Data Annotation} 
Before scaling LLM-based sentiment extraction to the entire dataset, we establish a validation set by randomly selecting 400 reviews for manual annotation to evaluate accuracy. Given the inherent challenges of manual labeling, such as high labor intensity, inconsistent standards, and subjective biases, we implement a structured annotation process. Specifically, four independent annotators manually label the selected reviews based on our predefined aspects. To assess annotation quality and consistency, we measure inter-annotator agreement and resolve discrepancies through a majority voting approach, ensuring the reliability and robustness of the final annotated labels. Table \ref{tab:examples} provides representative examples for each aspect, aiding in the evaluation and standardization of sentiment classification.

\subsection{Aspect-based Sentiment Analysis (ABSA)}
ABSA is a refined form of sentiment analysis that focuses on identifying and extracting opinions about specific attributes or aspects of a product or service, rather than assessing overall sentiment. This method enables a more nuanced understanding of user feedback by linking sentiments to particular service dimensions \cite{zhang2023survey}. In the context of this study, we leverage ABSA to extract sentiments associated with distinct five aspects of urgent care services from Google Maps reviews. Unlike traditional sentiment analysis, ABSA facilitates a more nuanced evaluation of patient experiences, thereby offering deeper insights into the specific areas that influence patient satisfaction in urgent care settings.

We design prompts to interact with LLMs to extract ABSA. A typical prompt consists of several components, including an instructional introduction, question prompt, text prompt, and expected output constraints. In addition, prior studies have demonstrated that LLMs exhibit strong few-shot learning capabilities \cite{brown2020language}, enabling them to generalize patterns from a few examples provided within the prompt. Research has also shown that providing examples in a prompt can improve accuracy and coherence, particularly in tasks requiring nuanced classification, such as ABSA. Therefore, we design the prompt by incorporating clear definitions, examples, and JSON output, aiming to optimize LLMs’ ability to generate reliable and contextually appropriate responses. The designed prompt is presented in Appendix\ref{sec:prompt}.

To further improve the reliability and consistency of aspect-based sentiment extraction, we adopt a structured prompt engineering strategy. Specifically, we provide the LLM with a system-level instruction that defines its role as an assistant for aspect-based sentiment analysis and explicitly constrains it to a predefined set of five healthcare-related aspects. The prompt further includes clear aspect definitions, representative examples, and a fixed JSON output schema to reduce ambiguity and prevent the generation of unsupported categories. This design aims to improve output stability, minimize hallucination, and enhance reproducibility, while allowing the LLM to flexibly interpret diverse and colloquial review texts.

\subsection{LLM Implementation and Performance Measure} 
\begin{figure*}[htbp]
\centering
\includegraphics[width=\textwidth]{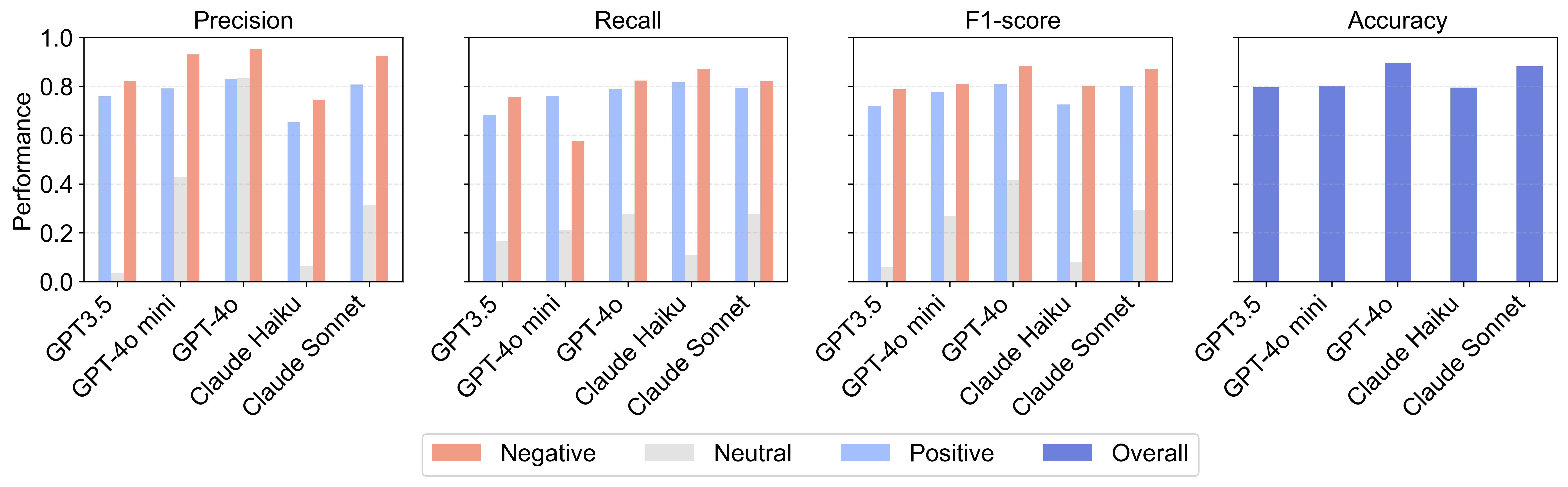}
\caption{Performance on the sentiment classification of candidate models.}
\label{fig:performance}
\end{figure*}

We evaluate the performance of five candidate LLMs: GPT-3.5, GPT-4o mini, GPT-4o, Claude-Haiku, and Claude-Sonnet on a testing dataset of 400 reviews using four key metrics: Precision, Recall, F1-score, and Accuracy. Precision indicates the percentage of true positive (TP) cases among predicted positives; Recall reflects the percentage of TP cases out of all actual positives. The F1-score provides a harmonic mean of Precision and Recall, and Accuracy measures overall classification correctness. The sentiment classification performance of each model is shown in Figure \ref{fig:performance}.  

\begin{table*}[htbp] 
\centering
\caption{Representative examples of ABSA from Google Maps reviews for urgent care.}
\label{tab:examples}
\begin{tabular}{p{0.6\textwidth}p{0.35\textwidth}} 
\hline
\textbf{Google Maps review [sic]} & \textbf{ABSA Classification} \\
\hline
We had a short wait time, got right in and the doctor was thorough, caring and quick to get us in and out and seen my family as a whole. Amazing visit thank-you next care at 501 N. Park Tucson, Az. &
\begin{tabular}[t]{@{}l@{}}
Interpersonal Factors: Positive \\
Operational Efficiency: Positive \\
Technical Quality: Positive \\
Facilities: Positive
\end{tabular} \\
\hline
Probably not worth the time. Staff was patient and listened well but weren't helpful. Charged me \$550 to give no answers and just told me to follow up with my regular doctor.&
\begin{tabular}[t]{@{}l@{}}
Interpersonal Factors: Positive \\
Operational Efficiency: Negative \\
Technical Quality: Negative \\
Finances: Negative
\end{tabular} \\
\hline
called at 4:50, they close at 5pm and the dr still saw me when i got there just before 5. very friendly and super knowledgeable. highly recommended its \$120 to see the dr without insurance &
\begin{tabular}[t]{@{}l@{}}
Operational Efficiency: Positive \\
Technical Quality: Positive \\
Finances: Neutral \\
Facilities: Positive
\end{tabular} \\
\hline
ABSOLUTELY TERRIBLE!! I called at 11:51 and nobody answered so I just decided to go anyways. I get there at 12:26 (the wind is blowing crazy at this point) and the doors are locked. Thinking I went to the wrong door I decided to walk around the building for another door. When I reached the first door I went to, there was now a HAND WRITTEN sign up saying that they are closed until 1:30.... &
\begin{tabular}[t]{@{}l@{}}
Interpersonal Factors: Negative \\
Operational Efficiency: Negative \\
Facilities: Negative
\end{tabular} \\
\hline
Was admitted from heart clinic bed not ready and you know the hospital wasn't full people talking temperature talking about going out and other things not related to job while standing in line they put me in ER people with no mask cussing hitting on doors lady sitting in wheelchair needs restroom no one helps so she pist herself this is ridiculous I pay insurance may not be much but no one needs to be treated like this and they want their co pay please &
\begin{tabular}[t]{@{}l@{}}
Interpersonal Factors: Negative \\
Operational Efficiency: Negative \\
Finances: Negative \\
Facilities: Negative
\end{tabular} \\
\hline
\end{tabular}
\end{table*}

The performance evaluation is based on a sample of 400 reviews, but it is important to note that the dataset is flattened, as each review may include sentiment classifications for multiple aspects. As a result, the distribution of sentiment classes is highly imbalanced: there are 676 positive samples, 180 negative samples, and only 18 neutral samples. This imbalance likely contributes to the challenge of achieving high performance for the neutral class, as purely neutral comments in patient reviews are relatively rare. 

Figure \ref{fig:performance} reveals several key insights. All five tested LLMs demonstrate an overall accuracy of more than 80\%. Notably, GPT-4o and Claude-Sonnet achieve comparatively higher test accuracy along with a more balanced and higher F1 score across all classes. However, their cost is over ten times that of more cost-efficient models such as GPT-4o mini or Claude-Haiku. Although GPT-4o mini exhibits slightly lower performance, the trade-off is marginal and generally acceptable. Given the cost-effectiveness where GPT-4o mini is substantially more affordable and the large scale of our dataset, we ultimately choose GPT-4o mini for analysis.

\subsection{Statistical Models} 
\label{sec:sta}
To examine major factors influencing hospital ratings, we perform both marginal (Pearson correlation) and joint analyses (multiple linear regression) across hospitals, incorporating both the five hospital-level sentiment scores and additional, regional demographic and socioeconomic attributes associated with the hospitals’ geographical locations. The multiple linear regression in particular accounts for the strong confounding correlations amongst sentiment score aspects (Figure \ref{fig:correlations}) and identifies the strongest determinants of perceived quality of urgent care, using hospital ratings as its proxy.

\subsubsection{Regression Models} 

We formulate two multiple linear regression models for the adjusted analysis of sentiment aspect vs. review rating associations. For either model, the dependent variable is the average Google rating for each hospital $i$, denoted as${\ Rating}_{i}$, which quantifies overall patient satisfaction based on publicly available reviews. \textbf{Model 1} includes the five sentiment scores as predictor variables capturing five key aspects of patient experience---Interpersonal Factors, Operational Efficiency, Technical Quality, Finances, and Facilities. These scores are calculated as averages across reviews per hospital, reflecting collective patient sentiment. That is, let ${SentimentScore}_{ij}$ denote the average score for hospital $i$ and sentiment aspect $j$, \textbf{Model 1} is formulated as:

$${\ Rating}_{i} = \beta_{0} + \sum_{j = 1}^{5}{\beta_{j}SentimentScore}_{ij} + \varepsilon_{i}.$$

Here, $\beta_{0}$ is the intercept, $\beta_{j}$s represents the coefficients of interest for the sentiment scores (Interpersonal Factors, Operational Efficiency, Technical Quality, Finances, and Facilities), and $\varepsilon_{i}$ is the independent error term \cite{wooldridge2016introductory}. This fully adjusted regression thus accounts for the strong interdependence amongst the five sentiment aspects (Figure \ref{fig:correlations}), identifying the strongest ``determinant'' aspects that drive hospital
customer rating.

\textbf{Model 2} includes all sentiment aspects as included in \textbf{Model 1} but additionally adjusts for demographic and socioeconomic characteristics of the hospitals' geographical location as covariates. These are region-level metrics as linked to Census Block Groups (CBG) that the hospitals are located under, including population density, median income, rent-to-income ratio, GINI Index, household below poverty rate, no insurance rate, and unemployment rate. Adjusting these variables in \textbf{Model 2} allows us to additionally control for potential
contextual effects of hospitals' surrounding areas. Specifically, let ${SocioeconomicScore}_{ik}$ indicate region-level metric $k$ for hospital $i$ (seven total), \textbf{Model 2} is formulated as:

\begin{align*}
{\text{Rating}}_{i} &= \beta_{0} + \sum_{j = 1}^{5}{{\beta_{j}\text{SentimentScore}}_{ij}} \\
&\quad + \sum_{k = 1}^{7}{{\gamma_{k}\text{SocioeconomicScore}}_{ik} + \varepsilon_{i}}.
\end{align*}

Compared to \textbf{Model 1}, the additional coefficients $\gamma_{k}$ represent effects for socioeconomic factors. Prior to inclusion in the regression models, all demographic and socioeconomic variables underwent z-score normalization to address scaling differences and enhance comparability.

\subsubsection{Sensitivity Analysis}

\textbf{Interaction.} To examine potential regional variations in sentiment-rating relationships, we conduct interaction analyses within our regression framework. Following standard statistical practice for interaction modeling, we center the predictor variables by subtracting their respective means to minimize multicollinearity effects. We include three interaction terms: interpersonal factors × operational efficiency, interpersonal factors × population density, and operational efficiency × population density to address potential geographical heterogeneity in sentiment-satisfaction relationships. These population density-associated interaction terms enable us to test whether the association between key sentiment aspects and hospital ratings varies across different levels of urbanization, which could potentially reveal important contextual influences on patient evaluation of urgent care services.

\textbf{Data filtering.} To ensure reliable sentiment analysis, we restrict the dataset to hospitals with at least 10 reviews for each of the five aspects, yielding a total sample of 534 hospitals. This filtering strategy minimizes noise from sparse review data, enhancing the robustness of the sentiment scores. To evaluate the potential impact of our stringent filtering criteria on the regression outcomes, we conduct a sensitivity analysis by modifying the inclusion parameters. Specifically, we eliminate the minimum review threshold requirement for the finances aspect, which exhibits the lowest review availability among the five sentiment dimensions. This modification substantially increases our analytical sample from 534 to 716 hospitals (34\% increase). We subsequently fit a modified regression model using this expanded dataset, excluding finances while retaining the remaining four sentiment aspects (interpersonal factors, operational efficiency, technical quality, and facilities) as predictor variables, following the structure of \textbf{Model 2}. This approach enables us to systematically assess whether our primary findings remain consistent across different filtering specifications and sample sizes, thereby evaluating the robustness of our analytical framework and the stability of our coefficient estimates independent of specific filtering decisions.

\textbf{Multicollinearity.}  Despite the strong correlation among both the sentiment scores (Figure \ref{fig:correlations}) and across demographic and socioeconomic variables (e.g., median income and poverty rate), we reason it is important to include variables simultaneously in a fully adjusted model to delineate true determinants driving urgent care satisfaction. This approach is admissible in particular thanks to the larger geographical scales and sample sizes of our included hospitals (n=534). Quantitatively, we assess the impact of multicollinearity by computing the Variance Inflation Factor (VIF) for each independent variable to ensure the stability and interpretability of the regression estimates \cite{obrien2007caution}. We observe that, across the five variables, the largest VIF is Interpersonal Factors, suggesting the most moderate impact of multicollinearity in our data (Table \ref{tab:vif}). The model's explanatory power is evaluated using R-squared and adjusted R-squared metrics.

\begin{table}[htbp]
\centering
\caption{Variance Inflation Factors (VIF) for Independent Variables in Model 2}
\label{tab:vif}
\begin{tabular}{p{6cm}p{2cm}}
\hline
\textbf{Variables} & \textbf{VIF} \\
\hline
\textbf{Sentiment Scores} & \\
Interpersonal Factors & 7.982441 \\
Technical Quality & 4.193950 \\
Operational Efficiency & 5.192045 \\
Finances & 1.942998 \\
Facilities & 1.727500 \\
\textbf{Socioeconomic Factors} & \\
Population Density & 1.048605 \\
Median Income & 1.734290 \\
Rent-to-Income Ratio & 1.184145 \\
GINI Index & 1.154306 \\
Household Below Poverty Rate & 1.808886 \\
No Insurance Rate & 1.344027 \\
Unemployment Rate & 1.231215 \\
\hline
\end{tabular}
\end{table}

\section{Results}
The results analysis explores two research questions. For \textbf{RQ1}, we identify patterns in public sentiment across different aspects of urgent care services. We examine two key aspects: the variation in sentiment distribution across different aspects in DMV and FL, and the factors and aspects that have the greatest impact on patient sentiment. For \textbf{RQ2}, we evaluate which sentiment dimensions independently predict facility ratings after accounting for local socioeconomic and demographic context.

\subsection{Sentiment Distribution and Geospatial Patterns} 
\begin{figure*}[htbp] 
\centering
\includegraphics[width=\textwidth]{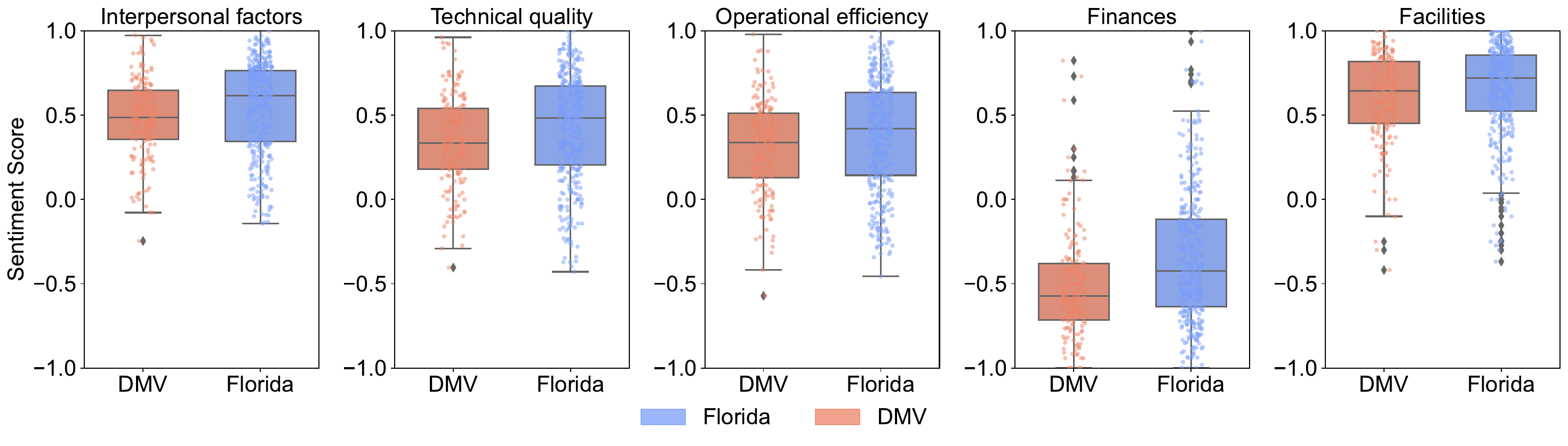}
\caption{Distribution of sentiment scores across five key aspects of patient experience by region.}
\label{fig:distribution}
\end{figure*}

Figure \ref{fig:distribution} shows the distribution of patient sentiment scores in the DMV region and Florida on the five dimensions of interpersonal factors, technical quality, operational efficiency, finance, and facility availability. The GPT-4o mini model is used to score sentiment, ranging from -1 (negative) to 1 (positive). Across all dimensions, Florida’s median sentiment is slightly higher than that of the DMV, especially in interpersonal factors and operational efficiency. The sentiment score of the financial dimension is generally low, and both regions have obvious negative emotions. The highest sentiment score is for facilities, indicating that patients have a good overall evaluation of the hospital's hardware conditions. Outliers indicate that there are extremely negative or positive comments on each dimension.

\begin{figure*}[htbp] 
\centering
\includegraphics[width=\textwidth]{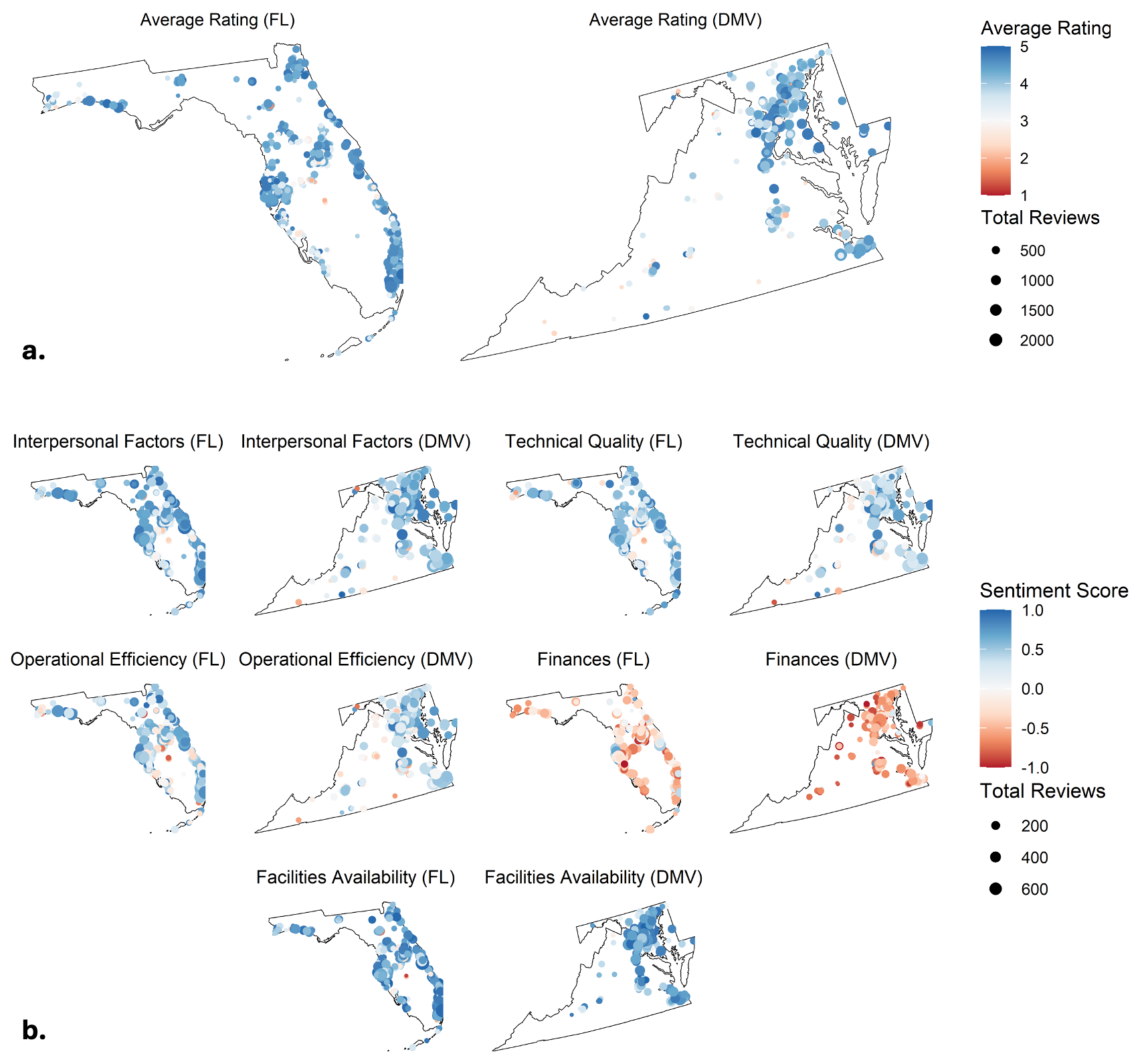}
\caption{Geospatial distribution among health centers. \textbf{a.} Average Rating Score in DMV and Florida. \textbf{b.} Sentiment Score among Aspects in DMV and Florida.}
\label{fig:geospatial}
\end{figure*}

Figure \ref{fig:geospatial} illustrates the geospatial distribution of urgent care centers in the DMV and Florida regions, highlighting spatial variations in review ratings and aspect-based sentiment scores across five dimensions. We overlay review data on state maps, where each dot represents an urgent care center; dot color indicates the average rating or sentiment score, and dot size represents the number of reviews. We include only urgent care POIs with at least 10 reviews. Our geospatial analysis of urgent care centers in Florida and the DMV region reveals notable trends in patient reviews and aspect-based sentiment scores. Overall, urgent care centers in Florida receive an average rating of 4.31 based on 172,597 reviews in 746 health centers while those in the DMV region have an average rating of 4.14 across 77,293 reviews in 378 health centers. These ratings suggest that patient experiences are generally positive in both regions. However, sentiment analysis across different aspects of care reveals important distinctions, particularly in financial satisfaction and operational efficiency. Financial concerns emerge as the most negatively rated aspect among the five key sentiment categories analyzed. In Florida, sentiment related to cost and billing has an average score of -0.34 across 11,701 reviews, while in the DMV region, financial sentiment is even lower at -0.53 from 5,002 reviews.

Other aspects of care are generally rated positively, though some differences emerge between regions. The sentiment of interpersonal factors, which reflects patient experiences with staff and provider interactions, is higher in Florida (0.59 across 89,357 reviews) than in the DMV region (0.49 across 40,631 reviews). Similarly, the sentiment of technical quality, which captures opinions on the care competency received, is higher in Florida (0.46 across 29,548 reviews) compared to the DMV (0.34 across 13,962 reviews).

Operational efficiency, including wait times and administrative processes, receives lower sentiment scores than other non-financial aspects, with Florida scoring 0.39 from 63,852 reviews and the DMV scoring 0.30 from 30,915 reviews. Meanwhile, facility-related sentiment, which measures perceptions of cleanliness, organization, and availability of medical equipment, received the highest ratings in both regions. Patients in Florida rated this aspect at 0.68 from 18,614 reviews, while those in the DMV region provided a similar rating of 0.61 from 7,755 reviews.

Overall, these findings indicate that urgent care centers in both regions are well-regarded for their interpersonal interactions, technical quality, and facilities, but concerns persist regarding financial transparency and operational efficiency. Florida's centers generally received higher sentiment scores than those in the DMV region, suggesting that patient experiences in Florida may be more favorable across multiple aspects of care.

\subsection{Statistical Analysis} 
\begin{figure*}[htbp] 
\centering
\includegraphics[width=\textwidth]{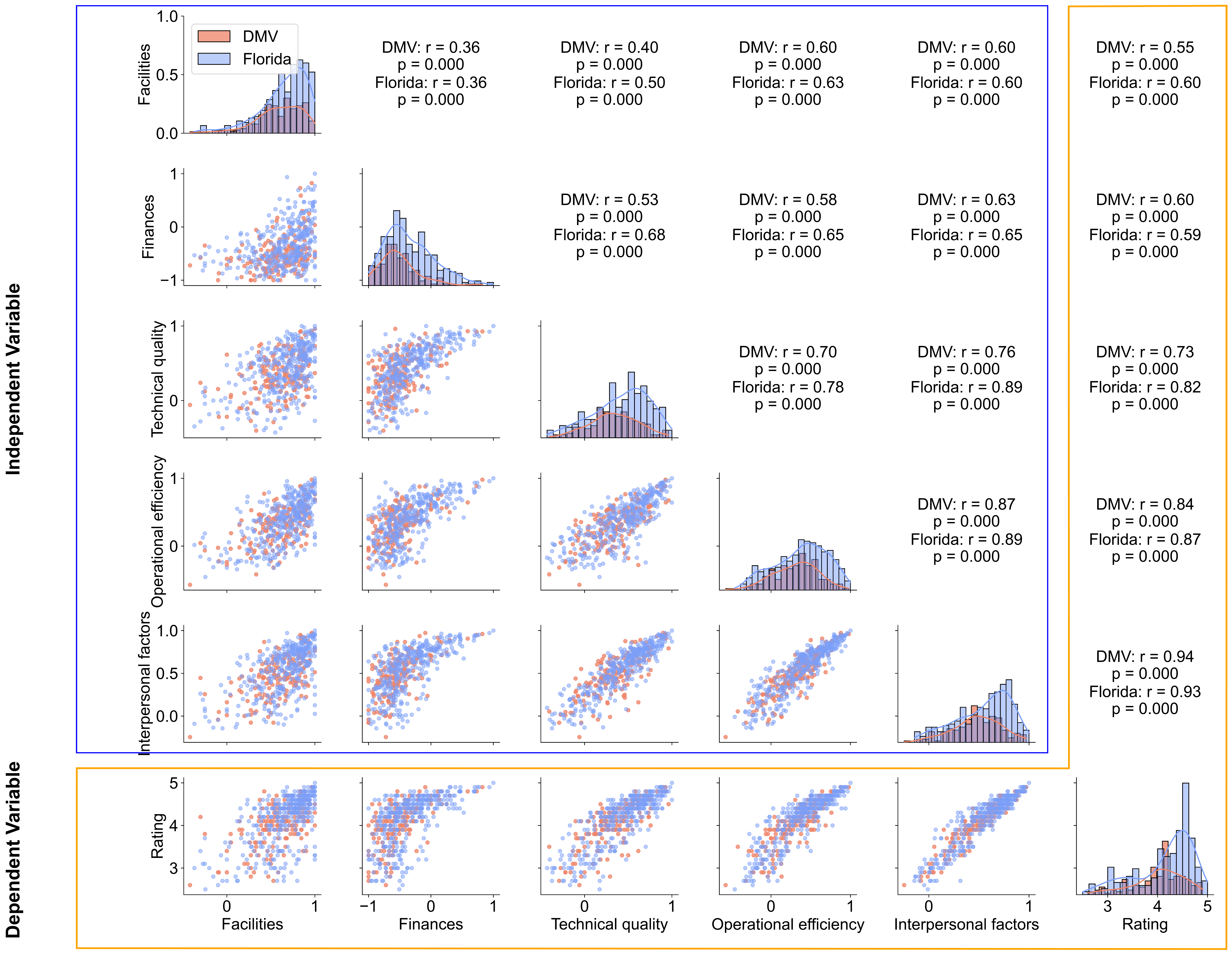}
\caption{Distribution (diagonal cells) and pairwise association (lower triangle), and pairwise correlations between service aspects and ratings for DMV and Florida regions.}
\label{fig:correlations}
\end{figure*}

All five sentiment scores are intercorrelated amongst each other and with average hospital ratings, suggesting different aspects of urgent care customer experience are typically related and jointly impact overall satisfaction (Figure \ref{fig:correlations}). We evaluate Pearson correlations between the five sentiment aspects (averaged per-hospital) and per-hospital average customer rating, stratified by region (DMV and Florida). The five aspects have significant positive intercorrelations (p < 0.0001 for all), ranging from moderate (facilities with finance, r=0.36 for both DMV and Florida) to strong dependencies (interpersonal factors with operational efficiency, DMV: r = 0.87; Florida: r = 0.89). 

The five aspects also have significant positive correlations with hospital ratings (p < 0.001 for all). Specifically, interpersonal factors exhibit the strongest correlation with hospital ratings in both DMV (r = 0.94) and Florida (r = 0.93), followed by operational efficiency (DMV: r = 0.84; Florida: r = 0.87) and technical quality (DMV: r = 0.73; Florida: r = 0.82). Finances (DMV: r = 0.60; Florida: r = 0.59) and facilities (DMV: r = 0.55; Florida: r = 0.60) show moderate correlations. This pattern is consistent in the two geographic regions, with Florida generally showing slightly stronger correlations for technical quality and operational efficiency. Together, our findings support that interconnected patient experience aspects jointly impact urgent care quality, as evaluated through public reviews and reproduced across two geographical locations.

We further delineate the individual impact of each urgent care experience aspect with a fully adjusted multiple linear regression analysis. Briefly (full details in Methods), we fit two linear regression models on per-hospital average review as the outcome. The first model includes all five sentiment scores as the predictors, thus delineating their individual (conditional) effects while accounting for the impact of other aspects. The second model additionally incorporates demographic and socioeconomic characteristics of CBGs that hospitals are located at as covariates (7 total; see Methods \ref{sec:sta} and Table \ref{tab:regression}) to further adjust for each hospital’s regional contexts.  Our conditional modeling approach allows us to determine, among inter-correlated per-hospital sentiment scores and area-level SES covariates, determinant factors that independently affect patients’ evaluation of the quality of their visit.

Our models establish consistent findings on significant predictors for visit satisfaction; in particular, interpersonal factors and operational efficiency are consistently the strongest determinants (Table \ref{tab:regression}). As an overall diagnostic, \textbf{Model 1} demonstrates the high explanatory power of sentiment scores on hospital rating (Adjusted R² = 0.875), which is only slightly improved with the additional CBG-level demographic and socioeconomic factors in \textbf{Model 2} (Adjusted R² = 0.877). Among the predictors, interpersonal factors emerge as the strongest predictor of hospital ratings across both models (\textbf{Model 1}: β = 1.684, p < 0.001; \textbf{Model 2}: β = 1.705, p < 0.001), followed by operational efficiency (\textbf{Model 1}: β = 0.303, p < 0.001; \textbf{Model 2}: β = 0.289, p < 0.001; Table \ref{tab:regression}). Notably, technical quality, finances, and facilities do not show statistically significant associations in either model, suggesting these are not major drivers of visit quality once interpersonal factors and operational efficiency are adjusted for. From \textbf{Model 2} results, population density is the only significant CBG-level factor, with more densely populated areas demonstrating higher ratings (β = 0.023, p = 0.009). Other SES variables do not show significant associations. Our results thus demonstrate interpersonal factors and operational efficiency as distinctive determinants shaping urgent care experience over other aspects (finance, facilities, technical quality), specifically addressing our \textbf{RQ2} (key aspects driving customer satisfaction with urgent care services). These agree with health economics research from other fields that highlight the importance of empathy, respect, and communication effectiveness in healthcare facility operations \cite{batbaatar2017determinants,derksen2013effectiveness}. Our findings, derived fully based on public review data and empowered by novel LLM analytics, thus join existing evidence in underscoring the importance of the ``human element'' in provider-patient interactions, in particular within the context of urgent care quality and satisfaction.

\begin{table*}[htbp]
\centering
\caption{OLS Regression Results for Predictors of Urgent Care Ratings}
\label{tab:regression}
\begin{tabular}{p{5.5cm}p{5.5cm}p{5.5cm}}
\hline
\textbf{Variables} & \textbf{Model 1: Sentiment Scores Only} & \textbf{Model 2: Sentiment Scores + Socioeconomic Factors} \\
\hline
\textbf{Sentiment Scores} & & \\
Interpersonal Factors & 1.684*** (0.091) & 1.705*** (0.091) \\
Technical Quality & 0.004 (0.056) & 0.004 (0.056) \\
Operational Efficiency & 0.303*** (0.062) & 0.289*** (0.061) \\
Finances & -0.055 (0.031) & -0.056 (0.031) \\
Facilities & 0.033 (0.041) & 0.036 (0.041) \\
\textbf{Socioeconomic Factors} & & \\
Population Density & - & 0.023** (0.009) \\
Median Income & - & 0.016 (0.011) \\
Rent-to-Income Ratio & - & 0.000 (0.009) \\
GINI Index & - & -0.011 (0.009) \\
Household Below Poverty Rate & - & 0.011 (0.011) \\
No Insurance Rate & - & 0.016 (0.010) \\
Unemployment Rate & - & 0.008 (0.009) \\
\textbf{Intercept} & 3.066*** (0.036) & 3.058*** (0.036) \\
\hline
\textbf{Model Fit} & & \\
R-squared & 0.876 & 0.880 \\
Adjusted R-squared & 0.875 & 0.877 \\
F-statistic & 748.9*** & 318.5*** \\
Observations & 534 & 534 \\
\hline
\multicolumn{3}{p{16.5cm}}{\centering\small Note: Standard errors in parentheses. Significance levels: ** p < 0.05, *** p < 0.001.} \\
\multicolumn{3}{p{16.5cm}}{\centering\small All socioeconomic variables in \textbf{Model 2} are standardized (mean = 0, standard deviation = 1).} \\
\end{tabular}
\end{table*}

Our findings are robust against regional variations and different model specifications through extensive sensitivity analyses (Methods \ref{sec:sta}). First, by additionally testing for interaction effects between interpersonal factors, operational efficiency, and region-level population density (i.e., those with significant coefficients), we observe no significant geographic heterogeneity in sentiment-rating relationships (interpersonal factors × population density: β = 0.080, p = 0.321; operational efficiency × population density: β = -0.087, p = 0.301).  Thus, key sentiment aspects consistently influence satisfaction regardless of urbanization levels. Second, to address the intercorrelations among sentiment scores, we evaluate multicollinearity via variance inflation factor (VIF) analysis. As shown in Table~\ref{tab:vif}, the VIF values for sentiment aspects are at most moderate and under thresholds deemed reasonable for large-scale studies with correlated predictors as discussed in O'Brien \cite{obrien2007caution}. Third, we evaluate the impact of our filtering criteria by comparing our primary regression results with an expanded sample analysis (Methods \ref{sec:sta}). This expanded analysis (n=716) excludes the finances aspect from our filtering criteria, which has the lowest availability in reviews and removes the most posts in the primary analyses (n=534). The findings are highly consistent with interpersonal factors (β = 1.64, p < 0.001) operational efficiency (β = 0.26, p < 0.001), and population density (β = 0.018, p = 0.024) still representing the strongest and only significant predictors. Our sensitivity analyses thus confirm the robustness of our primary findings, independent of specific model choices or regional characteristics.

We find that in \textbf{Model 2}, interpersonal factors show the largest coefficient of patient satisfaction (β = 1.705, p < 0.001). Additionally, operational efficiency is the second significant factor influencing patient satisfaction (β = 0.289, p < 0.001) towards their experiences with urgent care. On the contrary, despite technical quality showing a strong bivariate correlation with ratings (DMV: r = 0.73; Florida: r = 0.82), it demonstrates no significant independent predictive power in either multivariate model (\textbf{Model 2}: β = 0.004, p = 0.941). 

Even after adjusting for demographic and socioeconomic factors in \textbf{Model 2}, only population density was significantly associated with hospital ratings (β = 0.023, p = 0.009). The insignificance of other socioeconomic variables, including income, inequality measures, poverty rates, and insurance coverage, suggests that the relationship between patient satisfaction and service quality transcends socioeconomic boundaries within our sample.

\section{Discussion}

This study leverages large-scale Google Maps reviews and LLM-based aspect-based sentiment analysis to examine urgent care satisfaction across two geographically distinct regions. Guided by our research questions (RQ1 and RQ2), we interpret what patients value most in urgent care encounters and clarify which experience dimensions independently drive overall ratings after accounting for correlated aspects and area-level socioeconomic context.

\subsection{RQ1: Patterns across sentiment dimensions and their meaning}
Our RQ1 results highlight that online reviews contain rich, multi-dimensional signals of patient experience in urgent care. Beyond overall ratings, patients consistently comment on interpersonal interactions, operational efficiency, technical quality, finances, and facilities, reflecting both widely studied experience domains and additional topics often not captured in standardized surveys. Prior work has shown that unsolicited online narratives can complement traditional patient experience instruments and surface domains that matter to patients and caregivers~\cite{ranard2016yelp,kilaru2016edreviews}. In our setting, the co-occurrence of sentiments across aspects suggests that urgent care experiences are evaluated as an integrated encounter rather than isolated service components. Together, these findings provide a fine-grained characterization of how patients describe urgent care quality in real-world online reviews, supporting an urgent care-specific, multi-aspect framework for interpreting patient experience.

\subsection{RQ2: Determinants of satisfaction beyond socioeconomic context}
For RQ2, our adjusted models indicate that interpersonal factors and operational efficiency are the strongest independent predictors of overall satisfaction once correlated experience dimensions are jointly considered (Table \ref{tab:regression}). This pattern is consistent with the broader patient satisfaction literature emphasizing empathy, respect, and communication quality as central determinants of perceived care quality~\cite{batbaatar2017determinants,derksen2013effectiveness,manary2013patient}. Operational efficiency likely matters not only through actual throughput, but also through perceived waiting and information delivery, which have long been shown to shape satisfaction in acute care settings~\cite{thompson1996waiting,krishel1993info}. At the same time, aspects such as technical quality, finances, and facilities may be salient in narratives yet show limited independent explanatory power after adjustment, potentially reflecting shared variance with interpersonal and operational experiences, ceiling effects in perceived competence, or patients’ reliance on interpersonal cues when technical quality is difficult to assess directly. Overall, these results clarify which experience dimensions remain robust determinants of urgent care satisfaction under full adjustment, including area-level socioeconomic covariates, and help prioritize practical targets for quality improvement.

\subsection{Originality and methodological contribution}
While prior healthcare studies have mined topics and sentiment from online reviews~\cite{shah2021mining,ranard2016yelp,kilaru2016edreviews}, this study extends this line of research by providing a proof-of-concept demonstration of scalable, quantitative, and interpretable multi-aspect sentiment measurement in urgent care using recent LLM-based analytics. In particular, our prompt design constrains outputs to a predefined healthcare aspect schema with a fixed JSON format (Appendix\ref{sec:prompt}), improving reproducibility and interpretability relative to coarse-grained sentiment pipelines. In our evaluation across five candidate LLMs, newer models achieve higher and more balanced performance than earlier models, supporting cost-effective scaling to large review corpora.

\subsection{Practical implications for urgent care management (actionable recommendations)}
Our findings support specific, actionable steps that urgent care centers can implement to improve patient-perceived experience, especially along interpersonal and operational dimensions:
\begin{itemize}
    \item \textbf{Strengthen interpersonal interactions through structured communication practices.} Implement evidence-based communication skills training and reinforcement (e.g., greeting and explanation scripts, active listening behaviors, and de-escalation protocols), which can improve patient experience and satisfaction~\cite{brown1999communication,derksen2013effectiveness}. Given that interpersonal factors are the strongest determinant in our adjusted models, these interventions may yield substantial improvements in patient-perceived experience.
    \item \textbf{Reduce perceived waiting by improving transparency and information delivery.} In addition to operational process redesign, provide proactive updates on expected waiting time, triage rationale, and next steps at key touchpoints (front desk, triage, and discharge), as information delivery and perceived waiting have been shown to influence satisfaction in emergency care contexts~\cite{thompson1996waiting,krishel1993info}. Simple operational tools such as online pre-registration, queue visibility, and peak-hour staffing adjustments can further improve perceived efficiency.
    \item \textbf{Address recurrent financial frustration with standardized transparency steps.} Even when finances do not emerge as an independent driver after adjustment, consistently negative financial sentiment suggests a meaningful pain point in urgent care encounters. Centers can adopt standardized cost communication (e.g., up-front fee ranges for common visits, insurance verification scripts, and clear billing explanations at discharge), reducing surprise and frustration.
    \item \textbf{Use LLM-based ABSA as a continuous quality monitoring instrument.} Similar to how online narratives have been used to complement traditional surveys~\cite{ranard2016yelp,kilaru2016edreviews}, healthcare managers can monitor aspect-level signals over time to identify recurring concerns (e.g., increases in waiting-related complaints) and evaluate the impact of targeted interventions.
\end{itemize}
Finally, consistent with prior work on patient satisfaction determinants~\cite{batbaatar2017determinants}, we note that interpersonal and operational experiences are multi-component constructs. More detailed analyses of their specific determinants (e.g., which interaction behaviors or workflow steps matter most within each aspect) and formal evaluation of intervention effectiveness are important directions for future research.

\section{Limitations}

Our study, while offering valuable insights into patient satisfaction in urgent care through crowdsourced data, has several limitations that contextualize our findings and guide future research.

\begin{enumerate}[leftmargin=*]

    \item \textbf{Selection Bias in Google Maps Reviews}  
    Google Maps reviews may not reflect the broader patient population. Online reviewers often skew younger, more tech-savvy, and higher socioeconomic status, and those with extreme experiences (positive or negative) are more likely to post. This self-selection bias could limit the generalizability of our findings, especially for older or less digitally engaged patients. Future work could combine these insights with representative surveys like HCAHPS to validate results and capture a wider range of perspectives.

    \item \textbf{Performance of LLM-based ABSA}  
    Our sentiment analysis depends on GPT-4o mini, chosen for its cost-effectiveness (over ten times cheaper than GPT-4o or Claude-Sonnet) while achieving precision and recall above 80\%. However, it struggled with the neutral sentiment class due to a severe class imbalance (only 18 neutral samples out of 400). This could affect the accuracy of our sentiment scores and downstream analyses. Future studies could improve this by expanding the validation set, testing domain-specific fine-tuning, or using newer models to boost precision, especially for neutral sentiments.

    \item \textbf{Geographical and Contextual Scope}  
    Limited to urgent care in the DMV and Florida regions as of September 2021, our findings may not apply to other states or healthcare settings (e.g., emergency departments). Regional differences in healthcare systems and patient expectations could alter satisfaction drivers. Expanding the framework nationwide and across diverse care types would test its versatility and broader relevance.

    \item \textbf{Development Challenges}  
    Creating our five-aspect framework involved consolidating overlapping concepts from existing models, a complex task requiring careful judgment. Manual annotation of 400 reviews was also labor-intensive, with potential for human error despite majority voting. These challenges highlight the need for streamlined annotation methods in future work.

    \item \textbf{Observational Design}  
    Our cross-sectional approach cannot prove causality. Unmeasured factors, such as staffing levels or local events (e.g., COVID-19), might influence both sentiment and ratings. Longitudinal or quasi-experimental studies (e.g., analyzing policy changes) could better establish causal links.

\end{enumerate}

Addressing these limitations will strengthen the reliability and applicability of LLM-based analysis in healthcare research.

\section{Conclusions}
We employ ABSA to investigate five key components of patient experience in healthcare systems and their relationships with customer ratings on Google Map reviews for urgent care services in the DMV and Florida regions. Our results show that interpersonal factors are the primary predictor of patient satisfaction, with operational efficiency also playing a significant role. Our findings can guide healthcare providers in enhancing patient satisfaction through improved interpersonal interactions and operational processes.

\appendix

We additionally report the performance of a RoBERTa baseline as supplementary material.
\begin{figure}[htbp]
  \centering
  \includegraphics[width=\columnwidth]{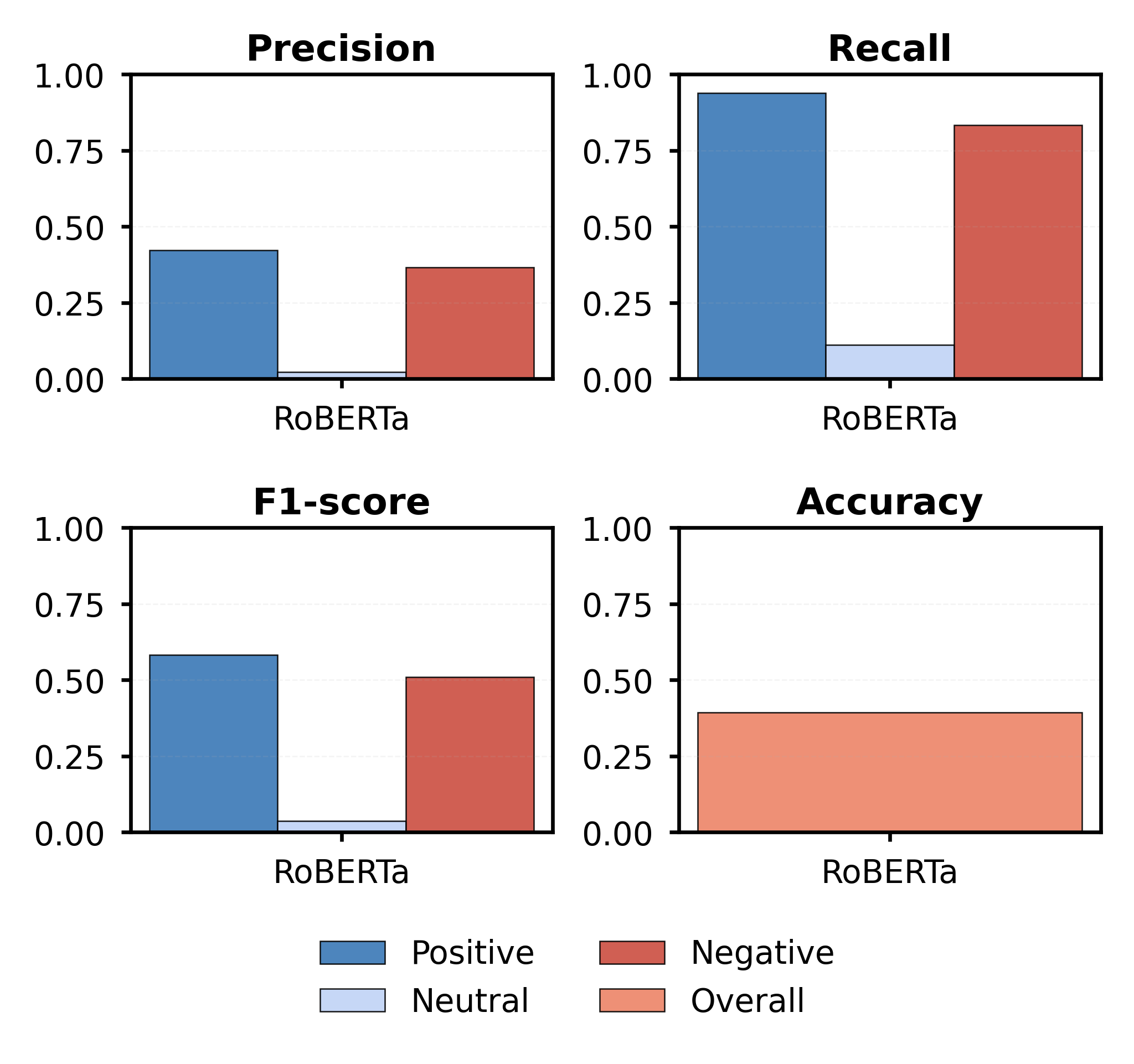}
  \caption{Multi-aspect sentiment analysis performance of the RoBERTa model across different metrics.}
  \label{fig:roberta_perf}
\end{figure}

\section{Prompt design}
\label{sec:prompt}
\begin{lstlisting}[language=Python]
system_prompt = "You are an assistant for aspect-based sentiment analysis. Only use the following aspects: Interpersonal Factors, Technical Quality, Operational Efficiency, Finances, Facilities/Availability. Do NOT create new aspects."

prompts = { 
    "intro": """### Sentiment Scoring of Patient Experience in Medical Care

Patient experience in medical care collected from Google Maps will be analyzed and categorized based on the following five key aspects. Each review will be classified as **positive, negative, or neutral** depending on the sentiment expressed in relation to these aspects:

**1. Interpersonal Factors**
- Features of the way in which providers interact personally with patients (e.g., concern, friendliness, courtesy, disrespect, rudeness).
- **Positive**: Empathy, kindness, active listening, respectful communication, friendliness, professionalism.
- **Negative**: Rudeness, indifference, lack of communication, dismissiveness, impatience, unprofessional behavior.
- **Neutral**: No clear sentiment, factual statements without strong emotional cues.

**2. Technical Quality**
- Competence of providers and adherence to high standards of diagnosis and treatment (e.g., thoroughness, accuracy, unnecessary risks, making mistakes).
- **Positive**: Accurate diagnosis, effective treatment, thorough examinations, expert medical knowledge, advanced technology use.
- **Negative**: Misdiagnosis, ineffective treatment, lack of expertise, medical errors, outdated procedures.
- **Neutral**: Descriptive statements without emotional judgment.

**3. Operational Efficiency**
- The results of medical care encounters (e.g., helpfulness of medical care providers in improving or maintaining health).
- **Positive**: Short waiting times, smooth appointment scheduling, efficient emergency response, streamlined administrative processes.
- **Negative**: Long wait times, disorganized processes, difficulty booking appointments, lack of coordination among staff.
- **Neutral**: Statements that report processes without clear sentiment.

**4. Finances**
- Factors involved in paying for medical services (e.g., reasonable costs, alternative payment arrangements, comprehensiveness of insurance coverage).
- **Positive**: Reasonable pricing, transparent billing, flexible payment options, adequate insurance coverage, financial assistance programs.
- **Negative**: High costs, hidden fees, billing errors, lack of insurance support, unexpected medical expenses.
- **Neutral**: Mentions of pricing or payment without emotional context.

**5. Facilities/Availability**
- Presence of medical care resources (e.g., enough hospital facilities and providers in the area).
- **Positive**: Clean and well-maintained facilities, modern medical equipment, comfortable waiting areas, availability of essential services (e.g., pharmacies, parking).
- **Negative**: Unhygienic conditions, outdated or broken equipment, lack of beds, overcrowding, poor accessibility for disabled individuals.
- **Neutral**: Observations about facilities without sentiment.

Each review will be assessed based on these five aspects to provide a structured sentiment analysis of patient experiences in healthcare settings.""",

    "question": """Can you identify the aspect-based sentiment (i.e., positive, neutral, negative) based on the patient experience stated in the following Google Maps review? Only use the following aspects: Interpersonal Factors, Technical Quality, Operational Efficiency, Finances, Facilities/Availability. Do NOT create new aspects. If the review does not mention an aspect, do not include it in the output.

### Example 1
Review: 'The doctor was very kind and took the time to explain everything to me in detail. The diagnosis was accurate, and I felt well cared for. The clinic was also clean and comfortable.'
Expected Output:
{
  "Interpersonal Factors": "positive",
  "Technical Quality": "positive",
  "Facilities/Availability": "positive"
}

### Example 2
Review: 'I had to wait for more than three hours even though I had an appointment. The staff was rude and unhelpful. Also, the bill had extra charges that were not explained to me.'
Expected Output:
{
  "Interpersonal Factors": "negative",
  "Operational Efficiency": "negative",
  "Finances": "negative"
}
""",

    "text": "The review content is: {review}\n",

    "output": """Please provide only a valid JSON response without Markdown code blocks, text descriptions, or explanations. 
The JSON structure must strictly follow this format:
{
  "Interpersonal Factors": "sentiment", 
  "Technical Quality": "sentiment", 
  "Operational Efficiency": "sentiment", 
  "Finances": "sentiment", 
  "Facilities/Availability": "sentiment"
}
Use only "positive", "negative", or "neutral" as sentiment values. 
If an aspect is not mentioned in the review, do not include it in the output.
If the review does not mention any of the defined aspects, return this exact JSON response: {"None": "None"}."""
}
\end{lstlisting}

\printbibliography

\end{document}